\documentclass{article}
\usepackage{ijcai23}
\usepackage{subfigure}

\usepackage{times}
\usepackage{soul}
\usepackage{url}
\usepackage[hidelinks]{hyperref}
\usepackage[utf8]{inputenc}
\usepackage[small]{caption}
\usepackage{graphicx}
\usepackage{amsmath}
\usepackage{amsthm}
\usepackage{booktabs}
\usepackage{algorithm}
\usepackage{algorithmic}
\usepackage[switch]{lineno}
\usepackage{stfloats}
\usepackage{multirow}
\usepackage{makecell}

\title{CPFES: Physical Fitness Evaluation Based on Canadian Agility and Movement Skill Assessment}

\author{
Pengcheng Dong$^1$\and
Xiaojin Mao$^2$\and
Lixia Fan$^2{}^,{}^3$\and
Wenbo Wan$^1$\And
Jiande Sun$^1$\textsuperscript{\rm †}
\affiliations
$^1$School of Information Science and Engineering, Shandong Normal University, Jinan, China\\
$^2$College of Physical Education, Shandong Normal University, Jinan, China\\
$^3$Education and Sports Bureau of Huaiyin District, Jinan, China
\emails
2022317067@stu.sdnu.edu.cn,
maoxiaojin0312@163.com,
\{fanlixia, wanwenbo\}@sdnu.edu.cn,
jiandesun@hotmail.com
}

\begin{document}

\maketitle
\renewcommand{\thefootnote}{}
\footnotetext[1]{\textsuperscript{†}The corresponding author.}

\begin{abstract}
In recent years, the assessment of fundamental movement skills integrated with physical education has focused on both teaching practice and the feasibility of assessment. The object of assessment has shifted from multiple ages to subdivided ages, while the content of assessment has changed from complex and time-consuming to concise and efficient. Therefore, we apply deep learning to physical fitness evaluation, we propose a system based on the Canadian Agility and Movement Skill Assessment (CAMSA) Physical Fitness Evaluation System (CPFES), which evaluates children's physical fitness based on CAMSA, and gives recommendations based on the scores obtained by CPFES to help children grow. We have designed a landmark detection module and a pose estimation module, and we have also designed a pose evaluation module for the CAMSA criteria that can effectively evaluate the actions of the child being tested. Our experimental results demonstrate the high accuracy of the proposed system.
\end{abstract}

\section{Introduction}
Recently, people pay more and more attention to the healthy growth of children, and positive sports activities play a crucial role in promoting their well-being. Not only can sports activities enhance their physical fitness, but they can also provide certain psychological benefits. Moreover, schools conduct physical fitness tests to evaluate the health and well-being of their students, ensuring their comprehensive physical and mental development.

The CAMSA\cite{LONGMUIR2017231} is one of the components of the Canadian Assessment of Physical Literacy (CAPL)\cite{longmuir2015canadian,longmuir2018canadian}. CAMSA restores the dynamic scenes of children and adolescents in their daily physical activities and physical exercises by connecting the seven motor skills of jumping on 2 feet, sliding from side to side, catching the ball, throwing the ball, step-hop,  1-footed hopping, and kicking the ball, evaluating the 8-12-year-old Fundamental Movement Skills (FMS)\cite{payne2017human} in children and adolescents, as well as the agility and comprehensive motor ability to execute combination skills and complex skills quickly. When children's physical fitness is lower, their completion will be lower than normal children. Therefore, the CAMSA is a method for screening children's physical fitness. If some children are found to be poor physical fitness, teachers and parents can be reminded to correct them in advance to avoid affecting their future growth.

With the development of deep learning, deep learning technology has been successfully applied to the fundemantal aspects of sports activities, such as \cite{song2021secure} and \cite{cust2019machine}, etc., making significant contributions to sports science. 
However, utilizing deep learning to evaluate the entire process of sports activities is still challenging.
Firstly, the duration of sports activities is usually long, and this computational effort is overwhelming if multiple videos of prolonged sports activity are used as input to the neural network.
Secondly, high-resolution videos are required for accurate evaluation of sports activities, which is difficult for neural networks to handle. 
In addition, wearable devices are developing rapidly.
At present, there are smart wearables with a variety of sensors like \cite{9395761}, which can not only measure physical data but also realize functions such as positioning and speed measurement. But many sports activities are difficult to evaluate purely by measuring physical data.

Currently, there are various applications and systems available for evaluating motor skills, such as \cite{quevedo2017assistance,maekawa2019naviarm}. These tools are not only used by clinical pediatricians and child development experts but are also increasingly being used as scientific research tools. Researchers utilize these tools to assess children's level of motor skill development and its correlation to sports participation, physical fitness, cognitive development, and academic achievement.

\begin{figure*}
    \centering
    \includegraphics[width=1.0\textwidth]{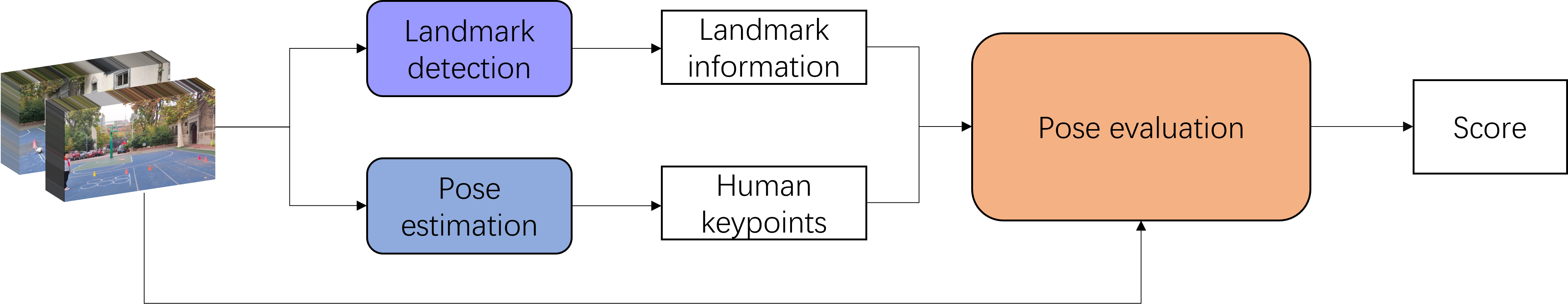}
    \caption{The structure of CPFES. CPFES consists of a landmark detection module, a pose estimation module, and a pose evaluation module.}
    \label{fig:my_label1}
\end{figure*}

In this paper, we propose a CAMSA-based Physical Fitness Evaluation System (CPFES) that scores human actions based on the relationship between human keypoints and the relationship between human keypoints and landmarks. The structure of CPFES is shown in Figure \ref{fig:my_label1}. The landmark detection module is used to detect the location of the landmarks of the test scene and confirm whether the number of landmarks matches the CAMSA. The pose estimation module is used to detect the pose of the child being tested and to obtain the keypoints of the human pose. The pose evaluation module calculates the score using the information obtained from the above two modules.
We apply deep learning to physical fitness evaluation, where we use them to extract the landmark and the human pose to find the relationship between keypoints in the human body to obtain a score for CAMSA. Compared with the traditional evaluation method, our system implements evaluation by computer and deep learning, eliminating the existence of subjective scoring. Additionally, our evaluation system scores the same many times, which improves credibility. Our system is easy to operate and has a fast processing time. The test scene only needs to be set according to CAMSA.

\begin{figure}[H]
    \centering
    \includegraphics[scale=.7]{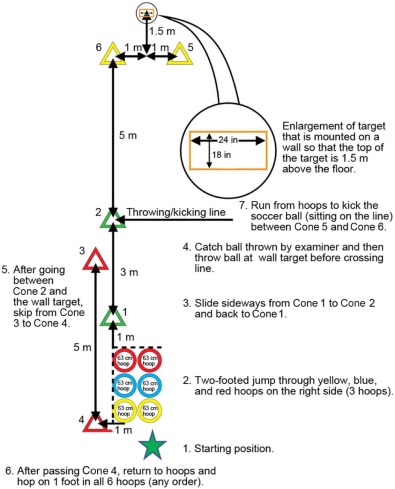}
    \caption{CAMSA course schematic. A video demonstration of the CAMSA is available on the website of the Canadian Assessment of Physical Literacy (\url{https://www.capl-ecsfp.ca/capl-training-videos}). CAMSA = Canadian Agility and Movement Skill Assessment. The figure is from \url{https://doi.org/10.1016/j.jshs.2015.11.004}. The action flow is marked in the figure.}
    \label{fig:my_label2}
\end{figure}

We summarize our contributions as follows:
\begin{itemize}
\item[$\bullet$] We developed CPFES to solve the shortcomings of traditional evaluation methods such as long evaluation time, strong evaluation subjectivity, and complicated operations. We apply deep learning to physical fitness evaluation and record videos through the CPFES system to get scores, which simplifies the physical fitness evaluation  process and improves efficiency. In addition, because of computer scoring, the scoring standards are fixed, which avoids subjective evaluation.
\item[$\bullet$] We design a pose evaluation module for CAMSA actions to evaluate the criteria for actions. Compared with traditional manual scoring, our pose evaluation module does not require human supervision and can achieve end-to-end output scores.
\item[$\bullet$] We design experiments to compare it with human evaluation, from which the accuracy of CPFES can be verified.
\end{itemize}

\section{Related work}

\subsection{Object detection.}

Object detection is one of the basic tasks in the field of computer vision, and object detection is already mature. Compared with traditional object detection algorithms, deep learning object detection can achieve end-to-end detection, and no longer requires multi-stage processing. There are two main categories of object detection, one-stage detection represented by YOLO\cite{7780460,8100173} and two-stage detection represented by R-CNN\cite{6909475,7410526,7485869}. Generally speaking, the two-stage detection method will be more accurate. However, in the current method\cite{8417976}, the difference between the two detection categories is already very small, but the one-stage detection method will be faster and more suitable for the detection of physical fitness activities. We use YOLOv5, a representative one-stage detection method, as the object detection algorithm in the CAMSA physical fitness evaluation system to detect the landmarks required by CAMSA.
\subsection{Human pose estimation}

Human pose estimation for images or videos is an important task of computer vision and serves as the basis for computers to understand human body language. Human pose estimation has many applications, such as for pose control and motion correction.
Significant progress has been made in human pose estimation. Human pose estimation is divided into bottom-up and top-down methods. The bottom-up approach\cite{8099978,8953507} first detects keypoints of all people in the image and then groups them into individuals. The top-down approach first detects the location of each person in the image and then predicts human keypoints for each person. The top-down approachs\cite{9156789} normalize human poses to approximately the same scale, so the best-performing methods so far are obtained by top-down methods. We use BalzePose\cite{bazarevsky2020blazepose}, a top-down human pose estimation method, as the pose estimation module of CPFES.

\begin{figure}
	\centering
	\subfigure[Front camera shot] {\includegraphics[scale=.3]{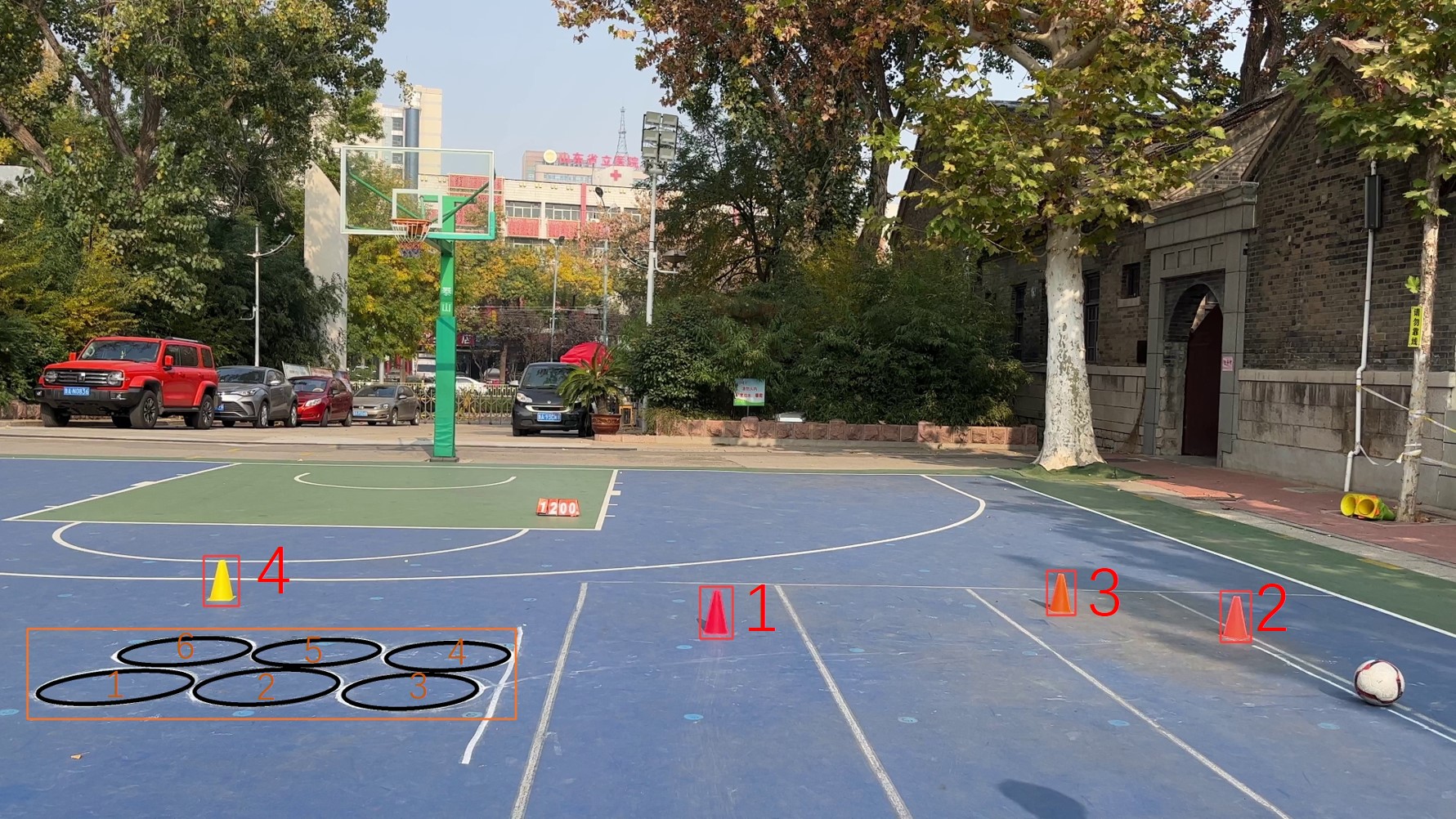}\label{3a}}
	\subfigure[Rear camera shot] {\includegraphics[scale=.3]{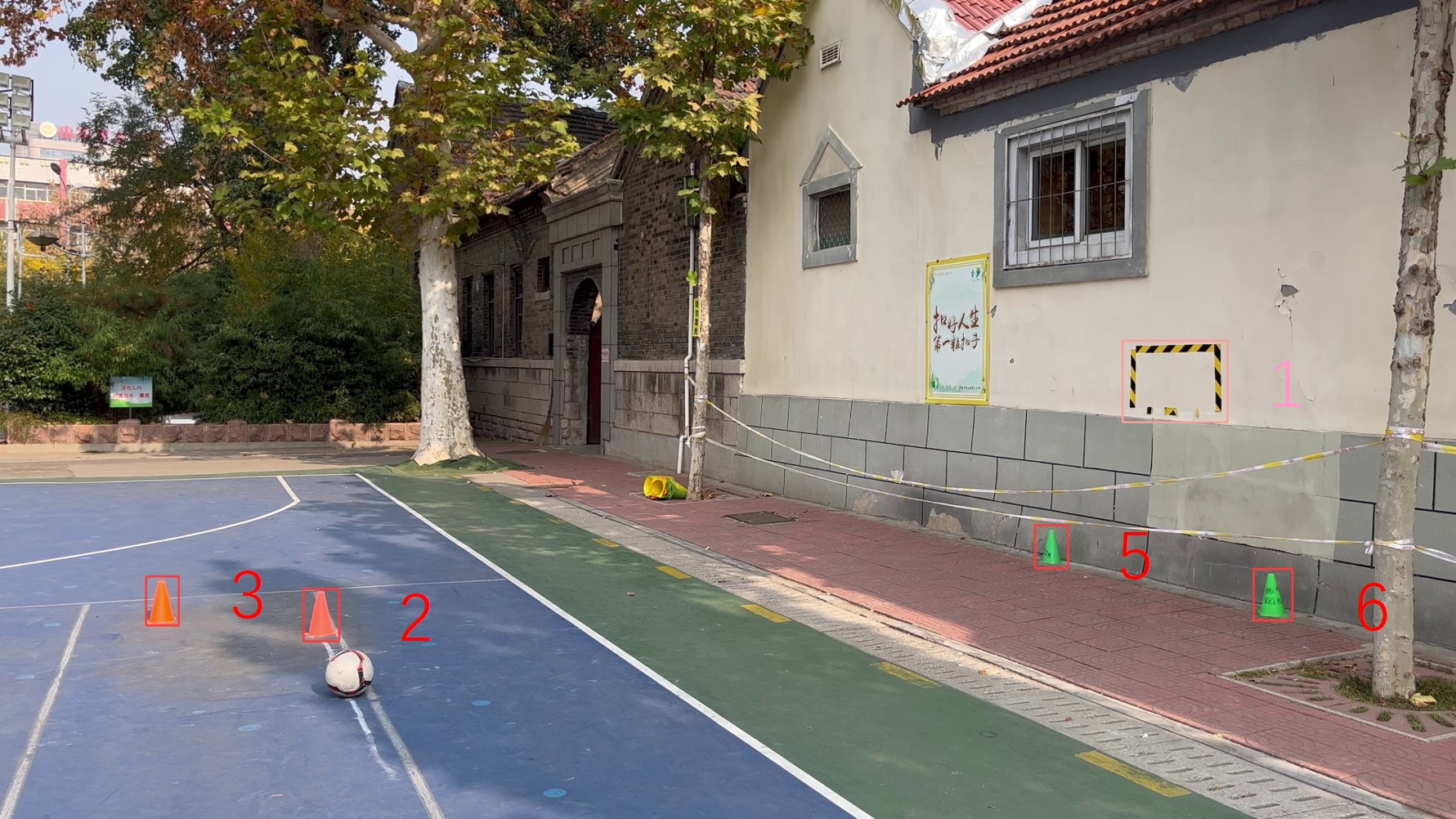}\label{3b}}
        \caption{CPFES test scene and landmark detection results. (a)The front camera records Action 1, Action 2, Action 3, Action 5, and Action 6. (b)The rear camera records Action 4 and Action 7. Action 1 is located at circular landmarks 1-3. Action 2 is located at triangle landmarks 1-2. Action 3 is between triangle landmarks 1-2, catch the ball. Action 4 is to throw the triangle landmark 2 places. Action 5 is located between triangle landmarks 3-4. Action 6 is located at circular landmarks 1-6. Action 7 kicks the ball at Triangle Landmark 2.}
	\label{fig:my_label3}
\end{figure}

\begin{figure}
    \centering
    \includegraphics[width=0.45\textwidth]{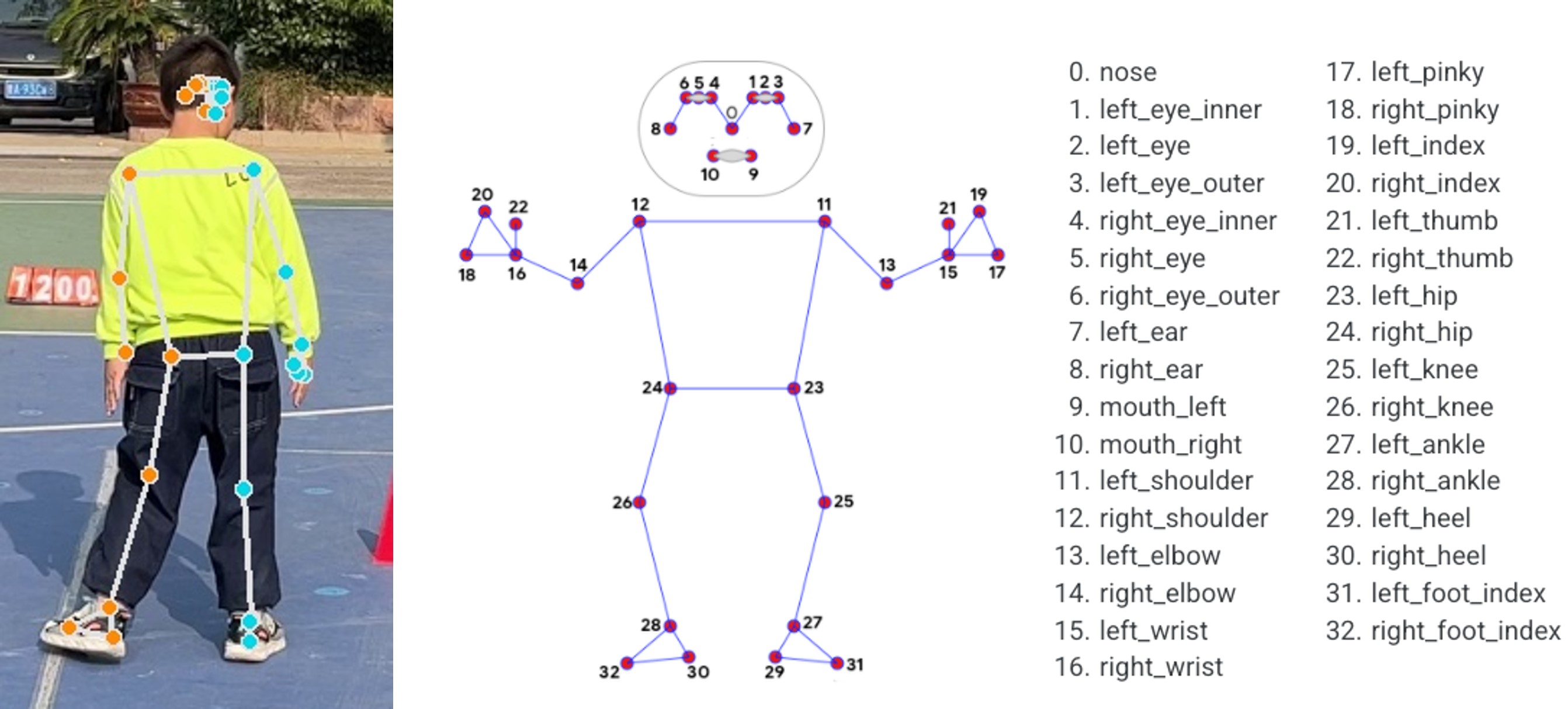}
    \caption{Pose estimation and BlazePose keypoints topology}
    \label{fig:my_label4}
\end{figure}

\subsection{Canadian Agility and Movement Skill Assessment}
Although there are many tools available for evaluating children's FMS internationally, CAMSA stands out as the first international closed-loop motor skill assessment tool based on a series of combined movements. By connecting seven motor skills, CAMSA is designed to restore the dynamic scenes of children and adolescents daily physical activities and exercises.

The CAMSA test scene is shown in Figure\ref{fig:my_label2}. The test children need to complete seven actions from the starting position, in the order of the landmarks, including jump on 2 feet, sliding from side to side, catching the ball, throwing the ball, step-hop,  1-footed hopping, and kicking the ball. The CAMSA score is divided into skill score and time score. The skill score and time score are 14 points for a total of 28 points. \cite{LONGMUIR2017231} gives detailed scoring rules.

\begin{table}[]
    \centering
    \begin{tabular}{ccc}
    \hline
        Completion time & Frame count& Score\\
        \hline
        \textless 14.0 & \textless 420 & 14\\
        14.0–14.9 & 420-449 & 13\\
        15.0–15.9 & 450-479 &12\\
        16.0–16.9 & 480-509 &11\\
        17.0–17.9 & 510-539 &10 \\
        18.0–18.9 & 540-569 &9\\
        19.0–19.9 & 570-599 &8\\
        20.0–20.9 & 600-629 &7\\
        21.0–21.9 & 630-659 &6\\
        22.0–23.9 & 660-719 &5\\
        24.0–25.9 & 720-779 &4 \\
        26.0–27.9 & 780-839 &3\\
        28.0–29.9 & 840-899 &2\\
        $\geq$ 30.0 & $\geq$900 & 1\\
        \hline
    \end{tabular}
    \caption{The completion time corresponds to the frame count. Since the video has a frame rate of 30 frames per second(FPS), we convert the time into frame count and use this to determine the time score.}
    \label{tab:my_label1}
\end{table}
\section{Method}
\subsection{Landmark detection moudle}
The CPFES test scene is planned according to the requirements of CAMSA, as shown in Figure \ref{3a}, where the landmarks required by CAMSA can be seen. 
Since the circular landmarks are compactly distributed and numerous, we group six circular landmarks into one type of landmark to facilitate detection. Therefore, the landmarks required by CAMSA can be grouped into one combined circular landmark, six triangular landmarks, and one rectangular landmark.
Our landmark detection module aims to accurately and quickly detect the landmarks required by CAMSA. To achieve this, we use YOLO v5 as the main method for landmark detection and output the coordinates of the landmarks in the image to determine their positions.
For circular landmarks, we use an edge detection algorithm\cite{4767851} to obtain their contour information. The algorithm first performs edge detection on the circular landmarks to obtain an image with edges marked and then applies dilation and erosion operations to the edge image to obtain the contour information of the landmarks.
We also added a detection mechanism to check whether the number of landmarks matches the number and location of landmarks in Figure \ref{fig:my_label2}, if not, the test scene needs to be adjusted. 
The results of landmark detection are shown in Figure \ref{3b}, which displays the detected positions of the landmarks in the CPFES test scene.

\subsection{Pose estimation module}
For the traditional manual evaluation method of physical fitness, multiple referees are required to evaluate the action and eliminate the subjectivity of the evaluation. 
We use deep learning for human pose estimation to improve evaluation efficiency and reduce subjectivity.
For scoring, the keypoints of the human body are crucial in expressing rich pose information and meeting the detailed requirements of scoring.
We use BalzePose for pose estimation, and the results of pose estimation are shown in Figure \ref{fig:my_label4}, which also displays the keypoints topology of BlazePose.
Blazepose has 33 keypoints. Compared with OpenPose\cite{8765346}, BlazePose can express more abundant pose information on the hands and feet, making it more conducive to judging the relative position of the human pose and landmarks.
Therefore, we choose BlazePose as the method for the pose estimation module in the CAMSA physical fitness evaluation system, due to its fast detection speed, high accuracy, and ability to express abundant pose information on the hands and feet. 

\begin{table*}[ht]
    \centering
    \begin{tabular}{llll}
        \hline
           Action & CAMSA criteria  & Point deduction  & CPFES criteria\\
        \hline
          Action1 & \makecell[l]{1.Three consecutive jumps, both feet\\ take off and land at the same time.\\2.No extra jumps without touching\\ the circle } & \makecell[l]{1.No jumping 3 times\\ in a row.\\2.Touch a circular landmark\\ or extra jump
         } & \makecell[l]{1.Keypoints 29-32 need to be within\\ the 1-3 circular landmark.\\2.Keypoints 27-28 have no vertical\\ movement.}\\
         \hline
          Action2 & \makecell[l]{3.Body and feet are aligned sideways\\ sliding in 1 direction.\\4.Body and feet aligned sideways\\ sliding in opposite direction.\\
         5.Touch cone when changing\\ directions after sliding} & \makecell[l]{3, 4.Cross the keypoints of\\ the legs during the round trip\\5.No touch landmarks}& \makecell[l]{3, 4. The line connecting keypoints 26\\ and 28 cannot intersect keypoints\\ 25 and 27.\\ 5. Keypoints 19 and 20 should touch\\ triangle landmarks 1 and 2.}\\
         \hline
          Action3 &\makecell[l]{6.Catches ball\\ (no drop or trap against body)} & \makecell[l]{6. Failure to catch the ball or\\ the ball touches the body} &\makecell[l]{6. Keypoints 19 and 20 are to\\ make contact with the ball.}\\
         \hline
          Action4 &\makecell[l]{7.Uses overhand throw to hit target.\\8.Transfers weight and rotates body\\ when throwing} & \makecell[l]{7.No hit the target\\8.no body turning to\\ assist in pitching} & \makecell[l]{7. The ball coincides with the position\\ of the rectangular landmark\\ 8. Keypoint 16 or 15 is relative to\\ keypoint 0 from back to front.}\\
         \hline
          Action5 &\makecell[l]{9. Correct step-hop foot pattern\\ when skipping\\10. Alternates arms and legs when\\ skipping, arms swinging for balance} & \makecell[l]{9.no running and jumping\\10.no arm swing}& \makecell[l]{9.Keypoints 27 and 28 appear to be\\ vertically displaced.\\10.The positions of keypoints 15 and\\ 16 alternate relative to keypoint 0.}\\
         \hline
         Action6 &\makecell[l]{11. Land in each hoop when\\ hopping on 1 foot\\12. Hops only once in each hoop\\ (no touching of hoops)} & \makecell[l]{11.Two feet inside a\\ circular landmark.\\12.Round landmarks only\\ jump once.}& \makecell[l]{11.Keypoints 27 and 28 must have\\ a position difference.\\12.The keypoints 27 or 28 cannot be\\ displaced by the vertical direction.}\\
         \hline
          Action7 &\makecell[l]{13. Smooth approach to kick ball\\ between cones\\ 14.Elongated stride on last stride\\ before kick impact with ball}& \makecell[l]{13. The ball is not between\\ landmarks 5 and 6.\\14.Kicking without leg\\ swing} & \makecell[l]{13.The position of the ball is between\\ triangle landmarks 5 and 6.\\14. The position of keypoints 26 and 28\\ will be from back to front from the\\ connection line of keypoints 25\\ and 27 when kicking the ball.}\\
         \hline
    \end{tabular}
    \caption{CAMSA criteria, points deductions, and CPFES criteria. CAMSA has a total of 14 points for the action score, and each serial number in the table corresponds to 1 point. We converted the CAMSA criteria into the CPFES criteria based on the point deduction items, which are specific to the keypoints of the human body and test scene landmarks that need to be detected for each action.}
    \label{tab:my_label2}
\end{table*}

\subsection{Pose evaluation module}

Our pose evaluation module uses the landmark information and pose information extracted by deep learning, and compares the relationship between different information through traditional algorithms, and then scores the actions of the childrens. The pose evaluation module consists of two parts.

The first part is the action-scoring component. We convert the CAMSA criteria and point deduction items for each action into the criteria suitable for CPFES, as shown in Table \ref{tab:my_label2}. CPFES evaluates the pose and landmark information based on the set standards for each action. For Action 1 and Action 6, we use the pointPolygonTest algorithm to detect whether the tester is stepping on the line in Action 1 or hopping on one foot in Action 6 by testing whether the keypoints of the foot are in the circular landmark. For Action 3, Action 4, and Action 7, we use the three-frame difference method to detect the ball. This method is highly adaptable to dynamic environments and insensitive to changes in scene light, making it suitable for our outdoor and changeable test scene. For Action 1, Action 5, and Action 6, we detect the vertical displacement of the keypoints of the ankle to determine whether there is repeated jumping.
The second part of the module is the time-scoring component, and the criteria for the time score are shown in Table \ref{tab:my_label1}. We detect the FPS and total number of frames in the video. We calculate the total frames taken to complete all actions of CAMSA and converting the total number of frames into time score.

\begin{table*}[ht]
    \centering
    \begin{tabular}{ccccccccccc}
    \hline
     Group & Referee  & Action1   & Action2 & Action3 & Action4 &Action5 & Action6 &Action7 & Time score &Sum score\\
     \hline
        \multirow[c]{5}{*}{1} & 1     &1.7	&2.0	&0.5	&1.3	&0.9	&0.7	&1.3	&3.5 &11.9\\
        ~ & 2     &1.8	&1.6	&0.3	&0.7	&0.8	&0.8	&1.2	&3.5 &10.7\\
        ~ & 3     &1.8	&2.0	&0.3	&0.8	&0.8	&0.7	&1.3	&3.5 &11.2\\
        ~ & ave-manual & 1.77 & 1.87  & 0.37 & 0.93 & 0.83 & 0.73 & 1.27 & 3.5 & 11.27\\
        ~ & CPFES   &1.7	&1.9	&0.4	&0.9	&0.8	&0.8	&1.2	&3.5 &\textbf{11.2}\\
    \hline
        \multirow[c]{5}{*}{2} & 1     &1.8	&2.0	&0.6	&1.1	&0.8	&0.7	&1.5	&4.0 &12.5\\
        ~ & 2     &1.9	&1.5	&0.5	&1.1	&0.4	&0.5	&1.5	&4.0 &11.4\\
        ~ & 3     &1.2	&2.1	&0.5	&1.0	&0.3	&0.2	&1.0	&4.0 &10.3\\
        ~ & ave-manual & 1.63 & 1.87  & 0.53 & 1.07 & 0.50 & 0.47 & 1.33 & 4.0 & 11.40\\
        ~ & CPFES   &1.9	&1.8	&0.6	&1.0	&0.7	&0.6	&1.6	&4.0 &\textbf{12.2}\\
    \hline
        \multirow[c]{5}{*}{3} &  1    &1.8	&1.0	&0.7	&1.0	&1.1	&1.8	&2.0	&5.6 &15.0\\
        ~ & 2     &1.9	&1.5	&0.7	&1.3	&1.3	&1.4	&1.7	&5.6 &15.4\\
        ~ & 3     &1.8	&1.5	&0.7	&0.8	&1.2	&1.1	&1.5	&5.6 &12.6\\
        ~ & ave-manual & 1.83 & 1.33  & 0.70 & 1.03 & 1.20 & 1.43 & 1.73 & 5.6 & 14.87\\
        ~ & CPFES  &1.8	&0.9	&0.8	&1.0	&0.9	&1.5	&1.8	&5.5 &\textbf{14.2}\\
        
        \hline
    \end{tabular}
    \caption{Different test groups, different manual scoring average score comparison table. Action 1-7 represents the average score of each group of the 7 actions of CAMSA, and the time score is the average time score of each group. The score with 1, 2, and 3 are manually scored, and the score with ours is the score obtained by our CPFES. The score with ave-manual is the average score of the manual score.}
    \label{tab:my_label3}
\end{table*}

\section{Experiments}

\subsection{Experiment details}

\begin{figure}[]
	\centering
	\subfigure[PR$-$curve] {\includegraphics[width=.2\textwidth]{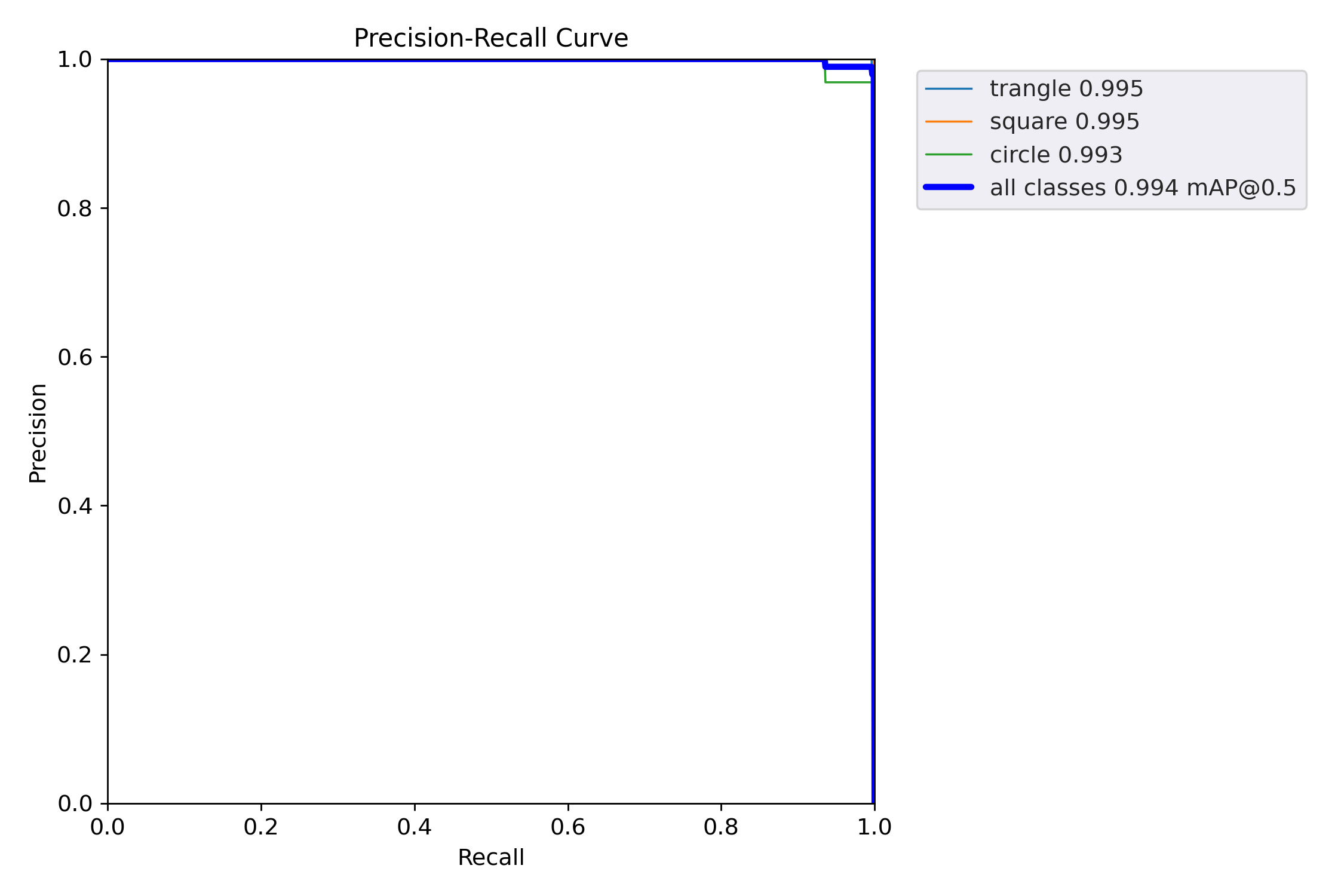}\label{5a}}
	\subfigure[F1$-$curve] {\includegraphics[width=.2\textwidth]{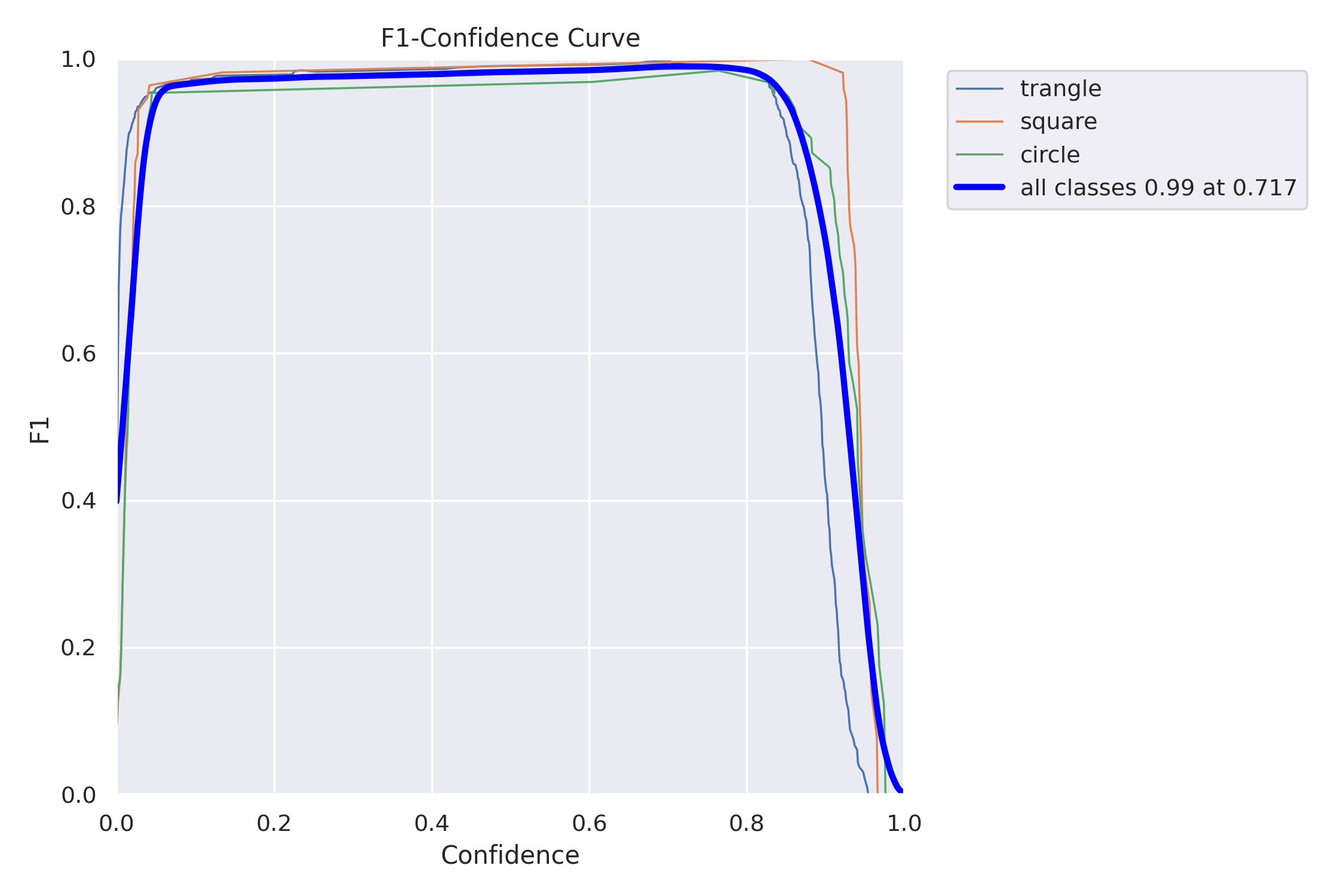}\label{5b}}
 
    \subfigure[P$-$curve] {\includegraphics[width=.2\textwidth]{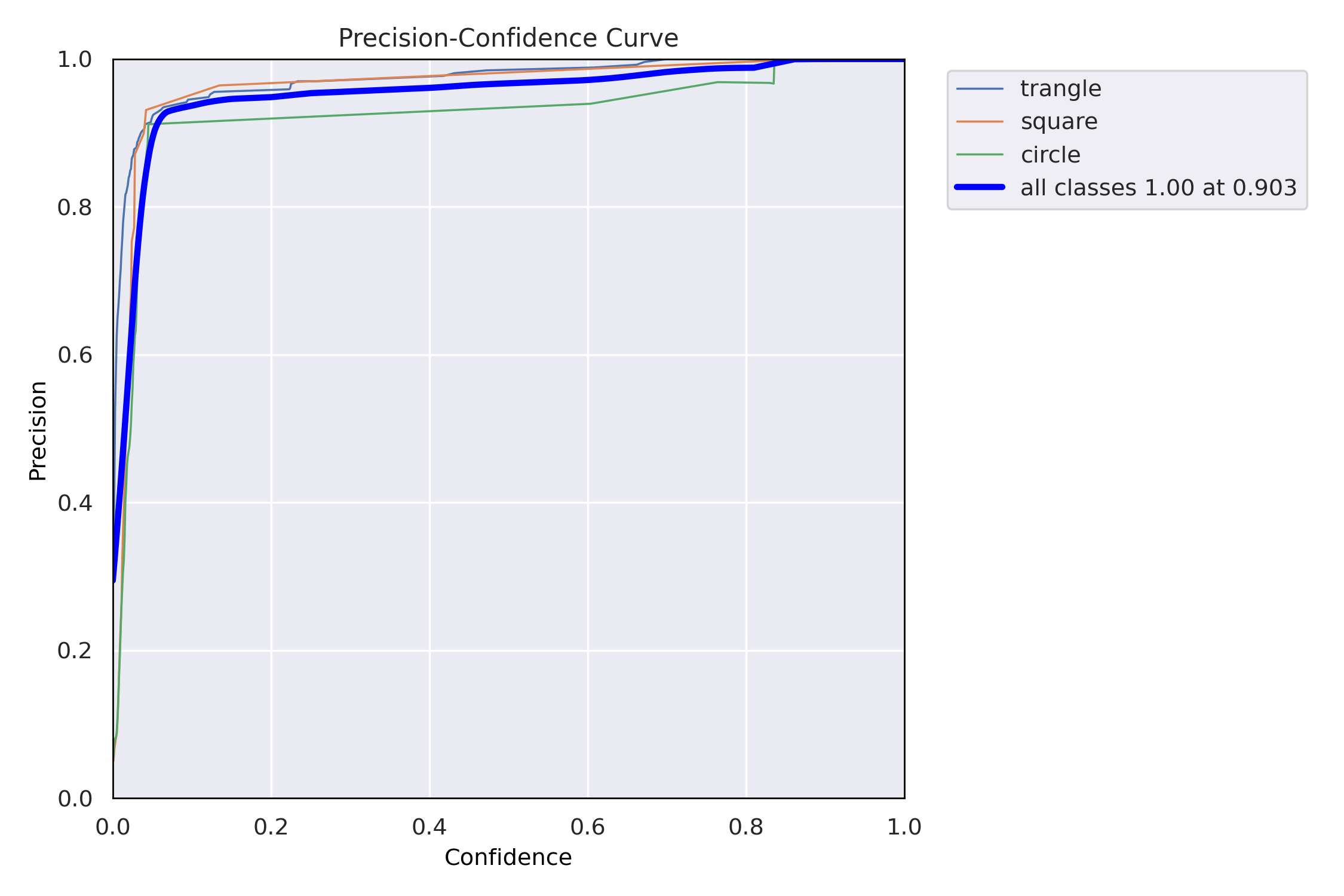}\label{5c}}
	\subfigure[R$-$curve] {\includegraphics[width=.2\textwidth]{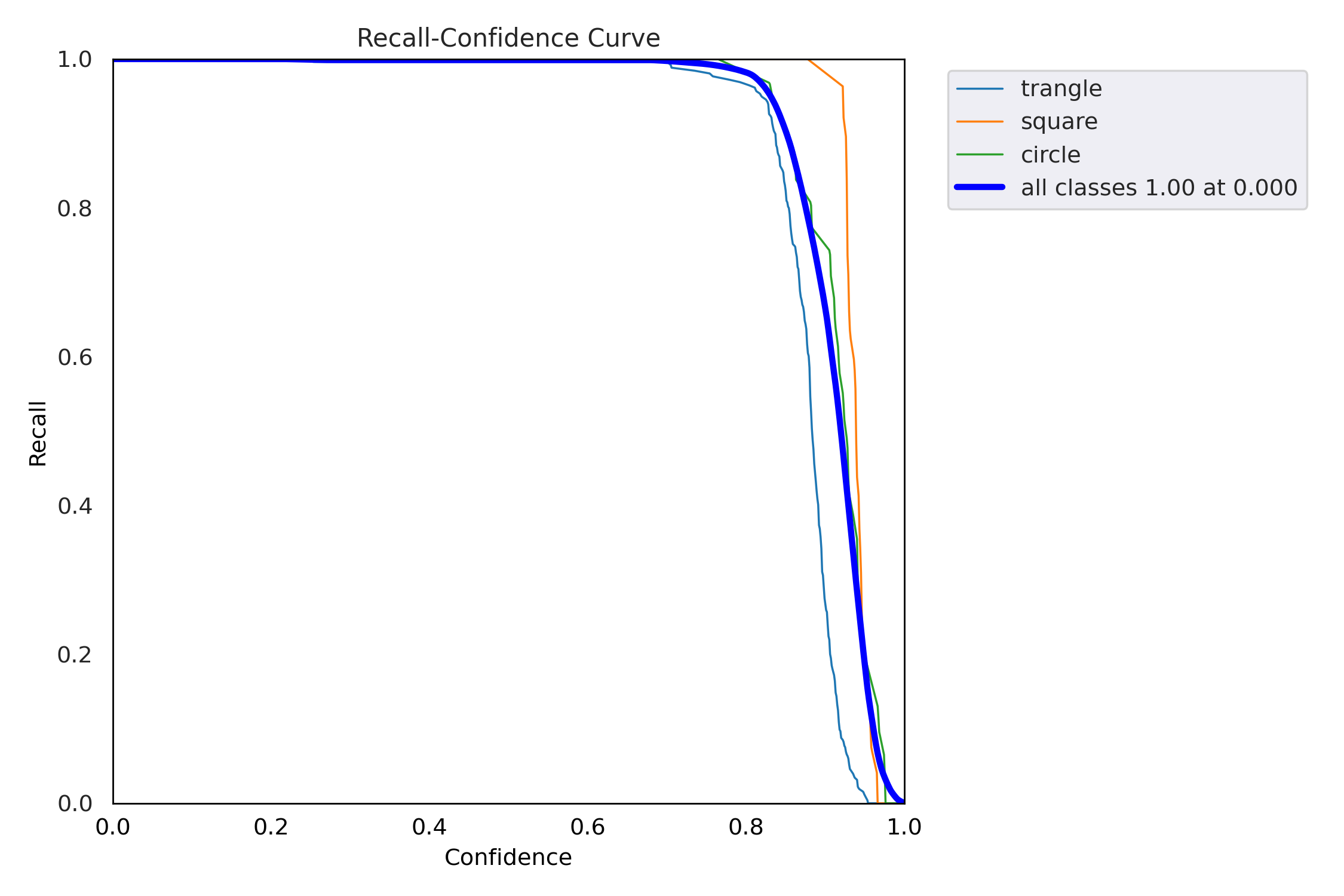}\label{5d}}
        \caption{Evaluation metrics chart.}
	\label{fig:my_label5}
\end{figure}

We recruited 30 children between the ages of 8 and 12 who met the requirements specified by CAMSA to test the accuracy of our evaluation system. The children were divided into three groups of 10 based on their grade levels, with a total of 10 girls and 20 boys. During the testing process, we recorded the physical fitness evaluation using two 4k resolution cameras at 30 FPS, while three individuals assisted with scoring, ball-throwing, and time recording. We collected two videos from the front and rear cameras for each child, resulting in a total of 60 videos. As each child took a different amount of time to complete the overall process, the length of the video was also different. To ensure the objectivity of our system, we had three individuals separately rate the videos to identify any potential influence of human subjective factors.

We use RTX1050ti, i8-8750H and 16G RAM to train our landmark detection, and collected 100 images from the CAMSA test scene as our dataset, which is used to train the landmark detection module to obtain landmark information, and the image is shown in Figure\ref{fig:my_label3}. We tested in different environments to ensure the robustness of our system. Our CPFES is also run on the same device to evaluate its performance. 
For the pose detection module, we utilized the BlazePose-based pose detection algorithm from Google Mediapipe to estimate human pose. By leveraging a pre-trained model, obtaining the pose of the human body is easily achievable.

\subsection{Experimental results of Landmark detection}

We use deep learning for landmark detection, and the evaluation metrics for the model are shown in Figure\ref{fig:my_label5}. The training was performed for 100 epochs, and it is obvious that using YOLOv5 as the landmark detection module can achieve well training results. In addition, we have  conducted multiple inference tests using 4k resolution test images, and the speed of inference for each test image can be maintained at 18ms.

\subsection{Experimental results of CPFES and analysis}

\begin{figure*}[ht]
	\centering
	\subfigure[] {
    \includegraphics[scale=.3]{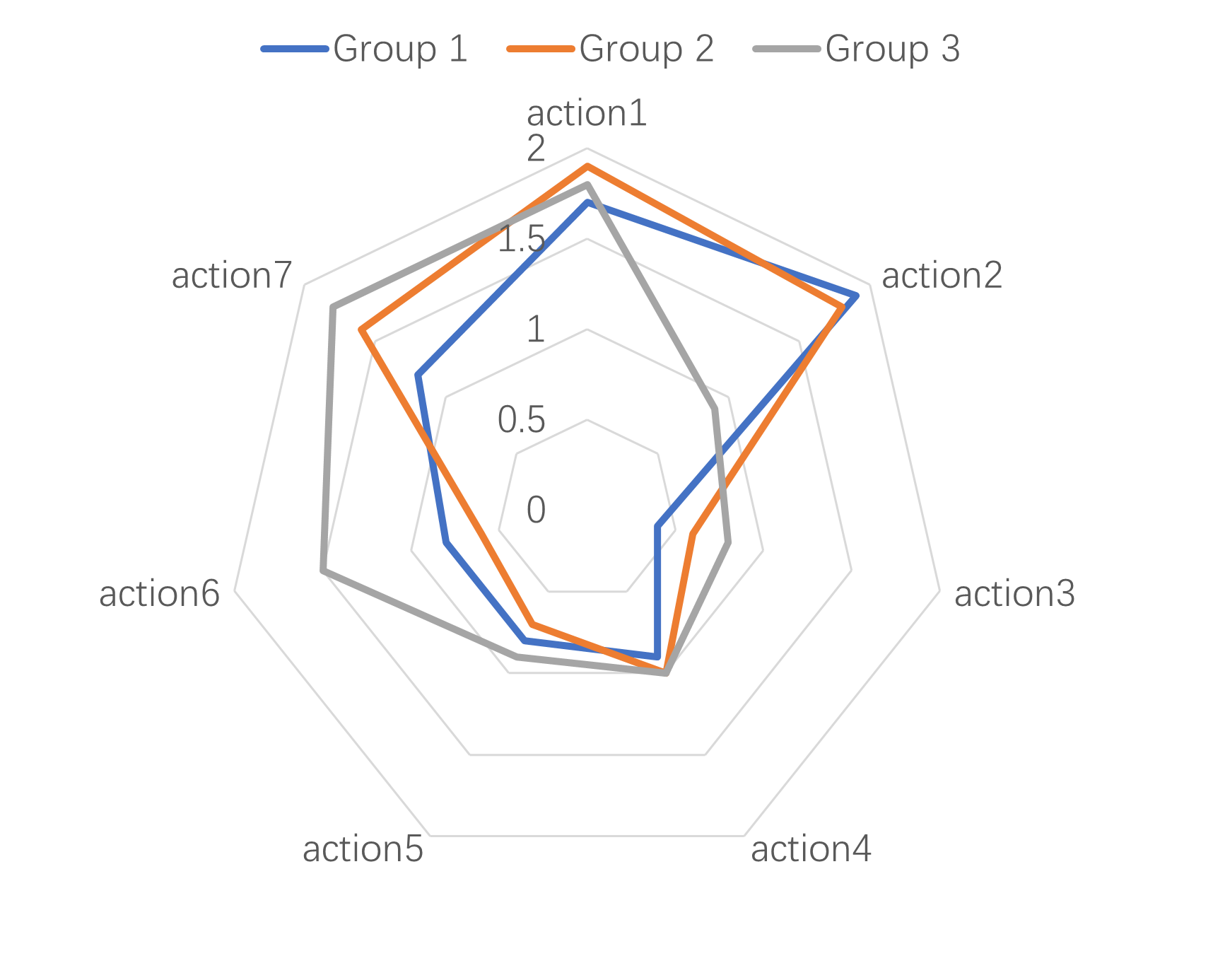}\label{6a}
    }
	\subfigure[] {
    \includegraphics[scale=.3]{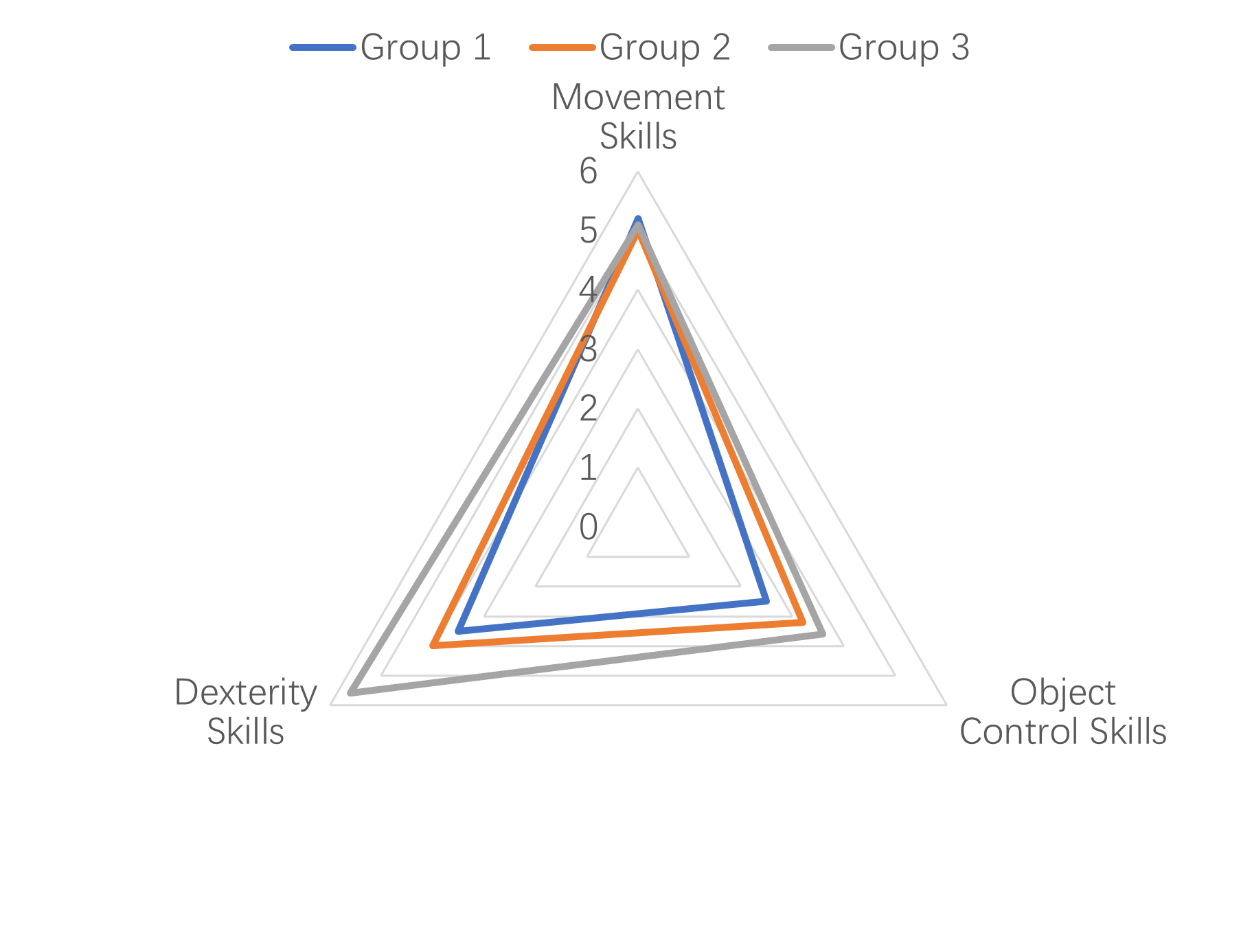}\label{6b}
    }
    \subfigure[] {\includegraphics[scale=.3]{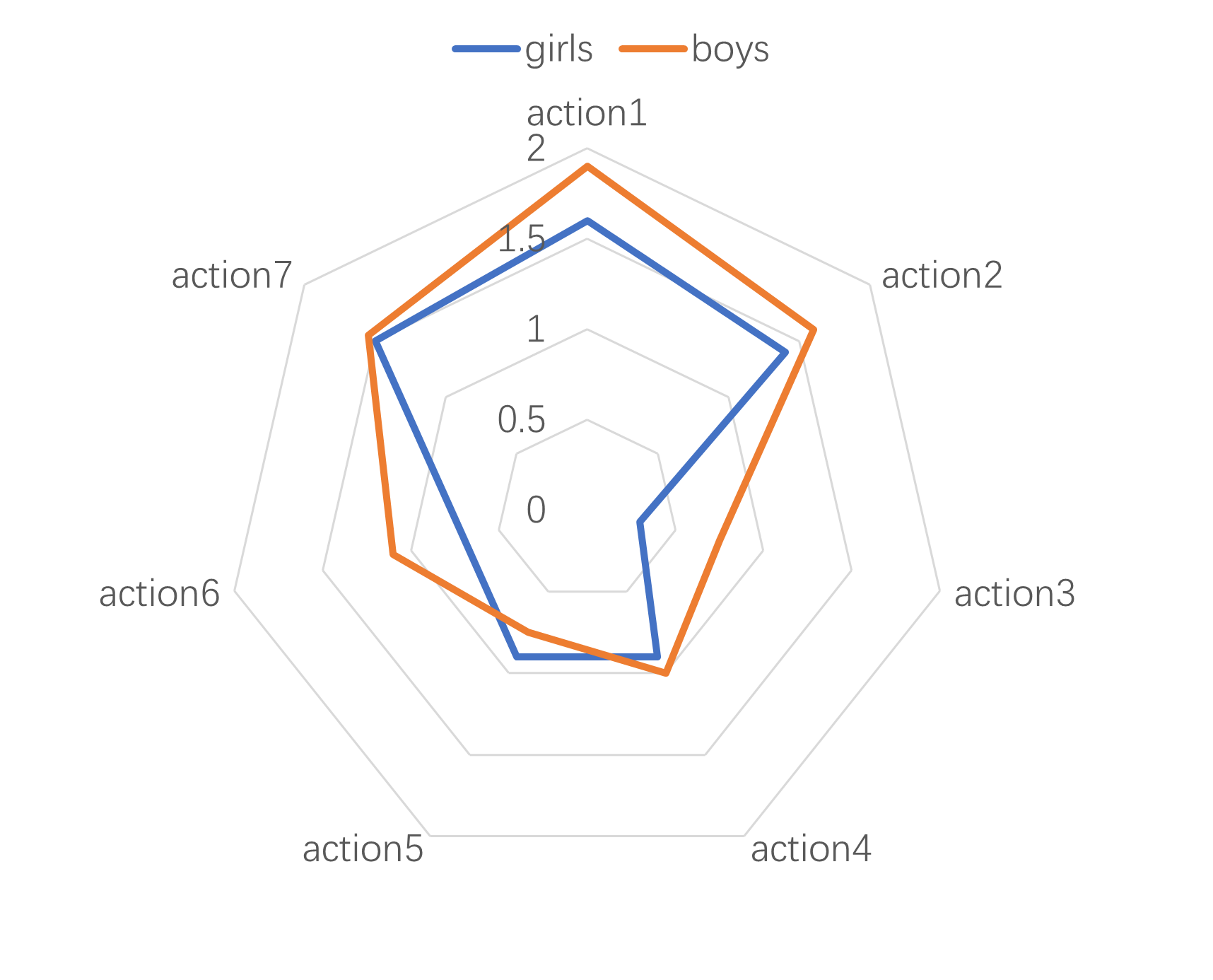}\label{6c}}
	\subfigure[] {\includegraphics[scale=.3]{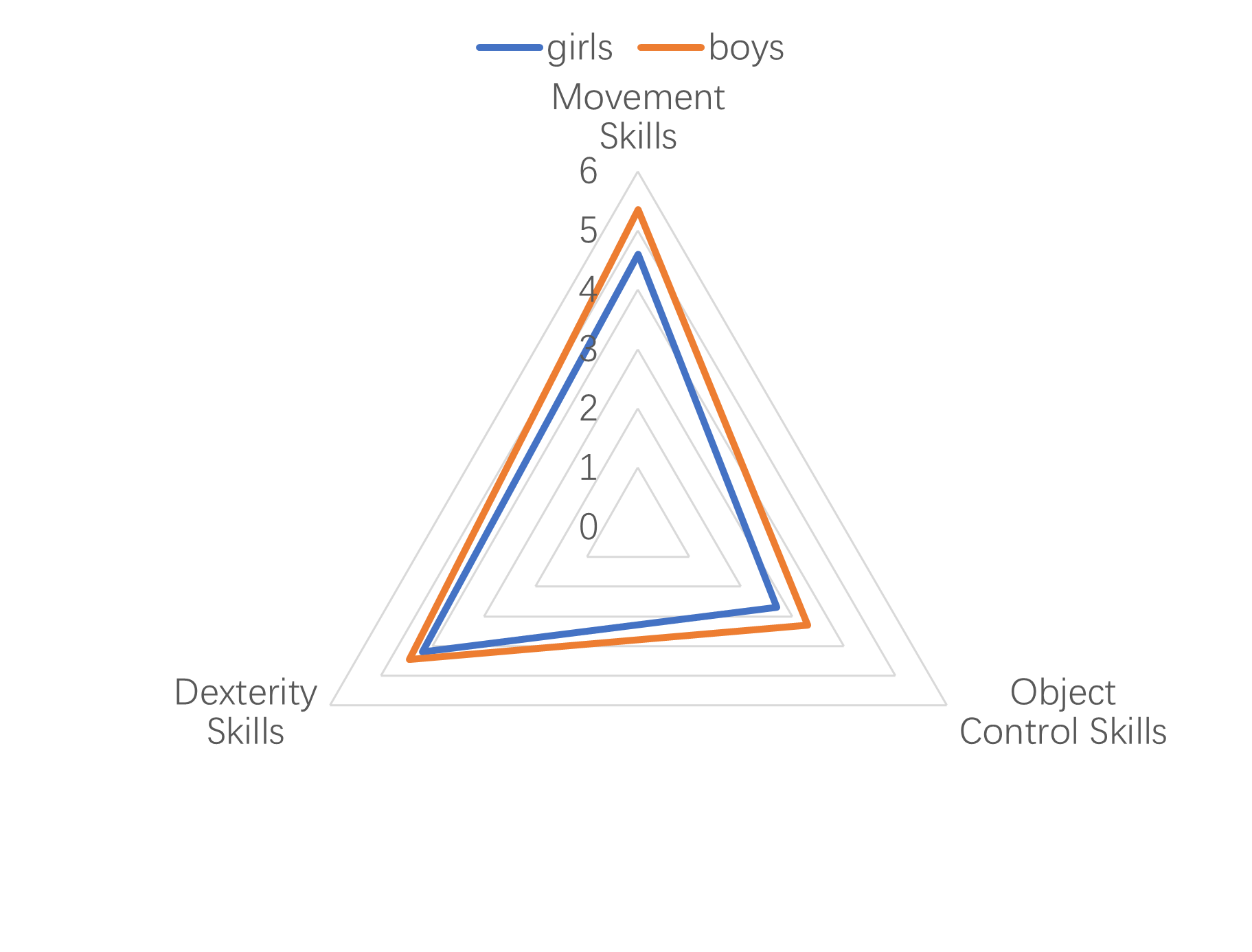}\label{6d}}
        \caption{Comparison of action scores and motor skills across different age groups and genders. (a) Comparison of action scores in different grades. (b) Comparison of motor skills in different grades. (c) Comparison of action scores of different genders. (d) Comparison of motor skills of different genders. }
	\label{fig:my_label6}
\end{figure*}
We scored 30 children according to the scoring criteria of CAMSA. We process the captured videos through CPFES and compare the obtained scores with human scores, as shown in Table \ref{tab:my_label3}, and we only use human score as our baseline.


Table \ref{tab:my_label3} shows that there is minimal difference in the total score between the CPFES and manual scores. Additionally, the time scores obtained by our CPFES are consistent with those of manual timing, with only a 0.1 difference for Group 3. Moreover, the scores obtained by CPFES for Actions 3, 4, and 6 are similar to those of the manual score, whether compared to different people or the average score of manual scoring.

For Action 1 of Group 2 and Action7 of Group 2, we can see that the score of the third referee is much lower than that of the first and second referees, but the score of CPFES is close to the scores of the first and second referee. In addition, the score of CPFES is also relatively close to the average score. So our system plays a role in weakening the subjective influence.

For Action 2, especially the comparison results of Group 3, our system's score is low. We observed the videos taken by different groups for comparison. We found that some testers in Group 3 only touched the apex of the triangle landmarks with their fingertips when they touched triangle landmarks 1 and 2 in Action 2. Although the pose estimation method we use can clearly express the pose information of the hand, it can not detect the fingertips, so the score of our system at this time will be low. This was not the case for the other two groups of testers overall.

For Action 5 of Group 3, the score between our system's score and the average manual score is 0.3. We analyze the pose information detected during Action 5. Since our camera is shooting from one side, we also estimate the pose of the tester from the side. We found that when performing pose estimation, human poses occasionally have false detection. For example, a certain keypoint will appear outside the human body, which is obviously unreasonable. This also has a great impact on the subsequent action scores, resulting in our low action scores.

Despite the differences between the CPFES score and the scores given by the three referees, we found that our CPFES score was closer to the average human score. This indicates that CPFES can help to mitigate the impact of subjective scoring.

CAMSA actions correspond to different motor skills, including jump on 2 feet, sliding from side to side strides, step-hop, and 1-foot hopping for movement skills, and catching, throwing, and kicking for object control skills. Time score represent dexterity skills. We divided the children into three groups based on their grades and compared the action scores among the groups. Furthermore, we classified the action scores based on motor skills and obtained a comparison chart of motor skills across the different groups. We also compared the action scores and motor skills between genders.

As shown in Figure\ref{5a}, the different groups performed similarly on action 4, while Groups 1 and 2 outperformed Group 3 on action 2, but Group 3 scored better on other actions. Figure\ref{5b} reveals that the three groups have similar movement skills, but Group 3 exhibits better object control and dexterity skills, which are positively correlated with grade. Based on the comparisons in the figures, we can conclude that the overall physical fitness of low-grade testers is weaker than that of high-grade testers, as demonstrated by the action scores and motor scores comparisons. Figure\ref{5c} shows that, except for action 5, girls scored lower than boys on the action scores. Figure\ref{5d} indicates that boys have stronger object control and movement skills than girls, while the dexterity skills of both genders are similar. In general, boys exhibit higher physical fitness than girls, except for dexterity skills where there is no significant difference.

Physical fitness levels vary significantly among children of different ages and genders. By comparing the radar charts, we can assign specific physical education classes or training programs to children of different grades or genders to improve their skills and scores. We also observed that boys need to improve their dexterity skills, while girls need to maintain consistent exercise to maintain good physical fitness. 
Furthermore, we suggest that Action 4 may not be the most suitable test for evaluating physical fitness in Chinese children, and other evaluation methods should be considered to gain a better understanding of children's physical fitness.

\section{Conclusion}
Our CPFES is composed of a landmark detection module, a pose estimation module, and a pose evaluation module. We used YOLO v5 as the foundation for the landmark detection module and BlazePose for the pose estimation module. We tailored our evaluation methods to the different actions of CAMSA and validated their accuracy through experiments.
Our system has a high degree of flexibility and can be adjusted using alternative methods to replace the landmark detection and pose estimation modules. We plan to reduce the complexity of the CPFES system and enhance detection accuracy by incorporating radar technology. Additionally, we will design a new hand network in the future to achieve more detailed keypoint detection. By collecting more test data, our system can compare and analyze the developmental status and physical fitness of children of different ages, genders, and regions, and explore the impact of various factors on children's development.

\section{Acknowledgments}
This work was supported by the General Program, Shandong Philosophy and Social Science Foundation of China (No.19CTYJ20).

\bibliographystyle{named}
\bibliography{ijcai23}

\end{document}